\documentclass[runningheads]{llncs}
\usepackage[T1]{fontenc}
\usepackage{graphicx,verbatim}
\usepackage{xurl} 
\usepackage{hyperref} 
\usepackage{amsmath} 

\usepackage{booktabs} 
\usepackage{multirow} 

\usepackage{amsmath,amssymb}
\usepackage{xcolor}
\usepackage{tikz}
\usetikzlibrary{positioning,fit,calc,arrows.meta}

\begin{document}
\title{CIV-DG: Conditional Instrumental Variables for Domain Generalization in Medical Imaging}
%

\author{Shaojin Bai\inst{1} \and
Yuting Su\inst{1} \and
Weizhi Nie\inst{1}\thanks{Corresponding author. Email: weizhinie@tju.edu.cn}}

\authorrunning{S. Bai et al.}

\institute{School of Electrical and Information Engineering, Tianjin University, Tianjin, China \\
\email{bsj072@tju.edu.cn, yutingsu@tju.edu.cn, weizhinie@tju.edu.cn}}

\maketitle             

\begin{abstract}

Cross-site generalizability in medical AI is fundamentally compromised by selection bias, a structural mechanism where patient demographics (e.g., age, severity) non-randomly dictate hospital assignment. Conventional Domain Generalization (DG) paradigms, which predominantly target image-level distribution shifts, fail to address the resulting spurious correlations between site-specific variations and diagnostic labels. To surmount this identifiability barrier, we propose \textbf{CIV-DG}, a causal framework that leverages \textbf{C}onditional \textbf{I}nstrumental \textbf{V}ariables to disentangle pathological semantics from scanner-induced artifacts. By relaxing the strict random assignment assumption of standard IV methods, CIV-DG accommodates complex clinical scenarios where hospital selection is endogenously driven by patient demographics. We instantiate this theory via a Deep Generalized Method of Moments (DeepGMM) architecture, employing a conditional critic to minimize moment violations and enforce instrument-error orthogonality within demographic strata. Extensive experiments on the Camelyon17 benchmark and large-scale Chest X-Ray datasets demonstrate that CIV-DG significantly outperforms leading baselines, validating the efficacy of conditional causal mechanisms in resolving structural confounding for robust medical AI.


\end{abstract}

\section{Introduction}
The clinical deployment of medical AI is frequently stalled by domain shift: models with expert-level internal performance fail to generalize across institutions~\cite{1,2,3}. While Domain Generalization (DG) typically attributes this to acquisition heterogeneity~\cite{4,5}(e.g., scanner physics, imaging protocols, staining), this pixel-centric view overlooks a more fundamental confounder—non-random patient selection into medical centers~\cite{6}. Unlike random noise, demographics and disease severity systematically dictate site assignment, inducing spurious site–label correlations~\cite{8,9,10} that feature alignment cannot resolve.

In observational multi-center studies, patient assignment is rarely random and is structurally entangled with demographics (selection bias)~\cite{11}. For example, patients with complex or rare pathologies are preferentially referred to tertiary hospitals, creating a dependency between patient characteristics and acquisition sites~\cite{12}. This confounding leads models to exploit site-specific artifacts (e.g., scanner noise or tag text) as shortcuts instead of learning causally invariant pathological semantics. Thus, DG methods that ignored this selection mechanism would break down if the demographic–site relationship were to shift~\cite{13,14}.

Instrumental Variable (IV) methods offer a principled route to causal identification by leveraging exogenous variation~\cite{16}. The acquisition site (hospital) is a natural instrument candidate: it affects the image generation process but should not directly cause the pathology~\cite{17}. However, standard IV requires instrument–confounder independence, which is often implausible in practice because hospital selection may be endogenous (e.g., driven by demographics and referral patterns), thereby undermining standard IV identification in clinical settings~\cite{18}.

Addressing this limitation, we introduce CIV-DG, a structural causal framework designed to generalize identification to scenarios plagued by selection bias. Departing from standard IV methods, CIV-DG leverages the theory of Conditional Instrumental Variables, which relaxes the strict independence assumption by positing that the instrument (hospital) is valid conditional on observed confounders (demographics). We instantiate this theory via a novel Deep Generalized Method of Moments (DeepGMM) architecture~\cite{19}. Specifically, instead of standard adversarial learning~\cite{5}, we employ a conditional moment critic to minimize the violation of conditional moment restrictions. This mathematically enforces the orthogonality between the instrument and the error term within demographic strata, thereby rigorously adjusting for confounding. As a result, CIV-DG effectively disentangles pathological semantics from site-specific artifacts~\cite{9}, ensuring that the learned representations remain invariant to both acquisition shifts and demographic selection biases.

Main contributions: 1) Causal Formalization: We introduce a novel perspective that identifies demographic confounding as a structural impediment to medical DG, formalizing cross-site generalization via Conditional Instrumental Variables (CIV) to enable identification where randomization fails. 2) Methodology: We propose CIV-DG, a structural causal framework implementing conditional moment restrictions with DeepGMM. By enforcing instrument–error orthogonality within demographic strata, CIV-DG addresses endogenous site assignment and disentangles pathology from site artifacts without paired data. 3) Clinical Validity: On Camelyon17 and Multi-Source Chest X-Ray, CIV-DG significantly outperforms state-of-the-art baselines, supporting fair, robust clinical AI.

\section{Method}

\begin{figure*}[!t]
    \centering
    \includegraphics[width=0.99\linewidth]{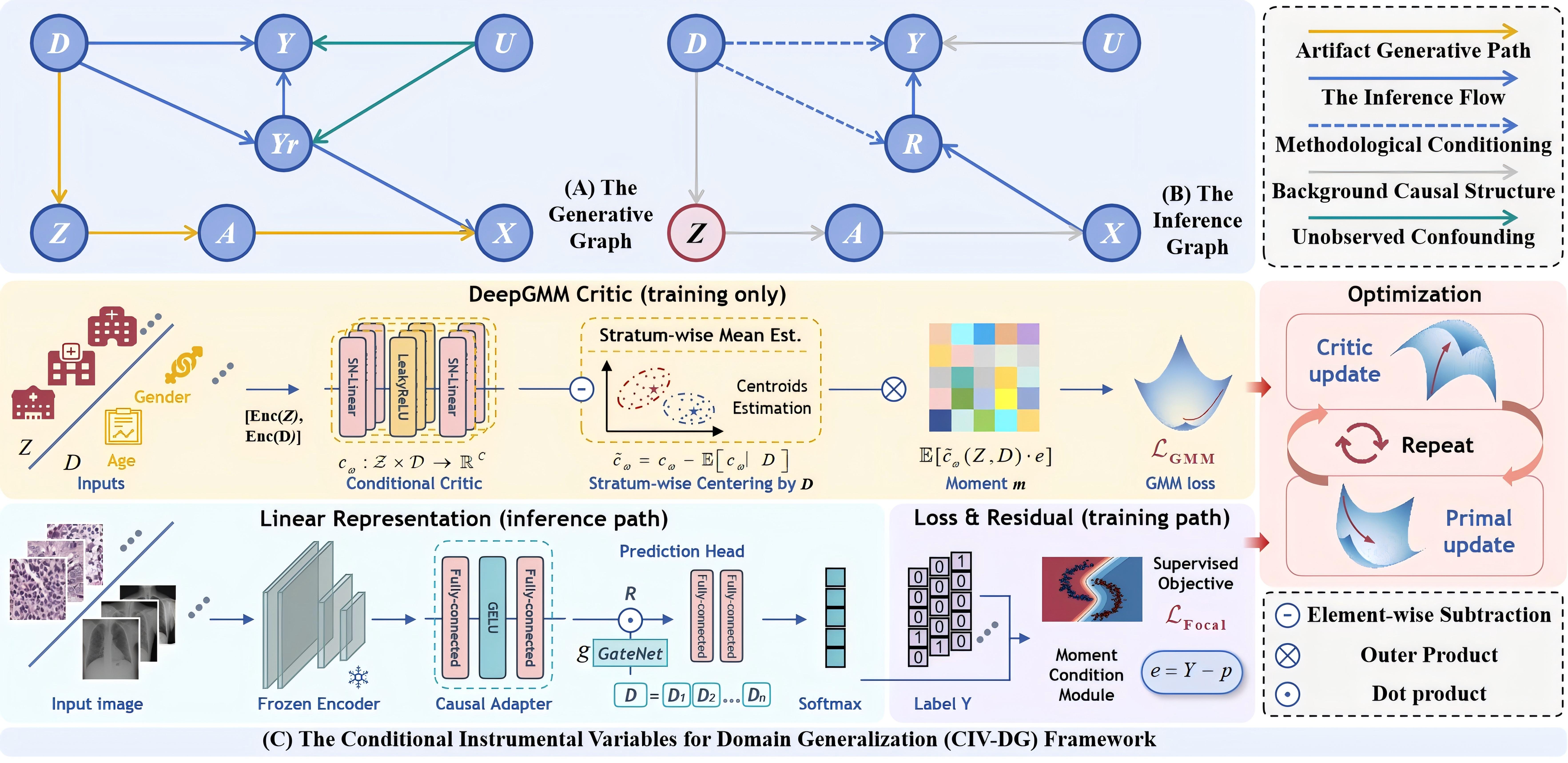}
    \caption{\textbf{Structural Causal Models (SCM) and the CIV-DG framework.} 
\textbf{(A)} The generative graph illustrating how hospital assignment $Z$ induces site-specific artifacts $A$ in images $X$. \textbf{(B)} Our inference graph. Dashed arrows from $D$ denote methodological conditioning rather than structural causal edges. \textbf{(C)} \textbf{The proposed CIV-DG framework.} It comprises the DeepGMM Critic (top) for enforcing conditional moment restrictions via stratum-wise estimation; the Linear Representation path (bottom) using a frozen encoder and Causal Adapter for prediction; and the Optimization module (right) depicting the alternating minimax updates between the critic and primal objectives.}
    \label{fig:1}
\end{figure*}

\subsection{Structural Causal Model (SCM)}

We formalize the data-generating process to clarify the link between the generative mechanism (Figure~\ref{fig:1} A) and our inference strategy (Figure~\ref{fig:1}B). Our goal is to disentangle clinically meaningful semantics from site-specific artifacts by translating causal assumptions into estimable moment conditions.

\noindent\textbf{Variables and Notation.} We consider a multi-site clinical imaging setting with observed variables $O := (X, Y, Z, D)$, comprising the image $X$, diagnostic label $Y$, hospital assignment $Z$, and demographics $D$. Additionally, we introduce the following latent mechanisms:
\begin{itemize}
\item $A$: site-specific imaging artifacts (e.g., scanner protocols, staining effects);
\item $Y_r$: the latent true clinical state underlying the image;
\item $U$: unobserved confounders (e.g., unmeasured severity or systematic annotation bias) that may jointly influence $Y_r$ and $Y$.
\end{itemize}

\noindent\textbf{Generative Graph.} Figure~\ref{fig:1} (A) encodes the data-generating process via the following structural equations, where $\epsilon_\cdot$ denote mutually independent exogenous noise terms:
\begin{align}
Z &:= g_Z(D, \epsilon_Z), \quad Y_r := g_r(D, U, \epsilon_r), \quad A := g_A(Z, \epsilon_A), \nonumber \\
X &:= g_X(Y_r, A, \epsilon_X), \quad Y := g_Y(Y_r, D, U, \epsilon_Y). \label{eq:scm_system}
\end{align}
Here, demographics $D$ drive selection bias ($D \to Z$) and influence the clinical state $Y_r$. Crucially, the image $X$ is a composite of the true signal $Y_r$ and site-induced artifacts $A$ ($Z \to A \to X$). Notably, $Z$ does not appear in the structural equation for $Y$, satisfying the exclusion restriction: hospital assignment influences the observed image $X$ (via artifacts), but has no causal influence on the diagnostic label $Y$.

\noindent\textbf{Inference Graph.} Our objective is to learn a representation $R := \phi(X)$ and a predictor $\hat{Y} := f(R, D)$, such that prediction relies on stable pathological features rather than spurious site correlations. Figure~\ref{fig:1} (B) depicts the inference graph, where dashed arrows from $D$ denote methodological conditioning. These indicate that $D$ is used for stratification to enforce conditional instrument validity (CIV) and may optionally serve as a predictor input for case-mix adjustment.

To achieve identification, we leverage hospital assignment $Z$ as a conditional instrument and posit the Conditional Instrument Validity (CIV) assumption: $ Z \perp U \mid D$. This assumption implies that, after controlling for patient demographics $D$, the site assignment $Z$ is effectively randomized with respect to unobserved confounders $U$. Consequently, within any demographic stratum, prediction errors should exhibit no systematic dependence on $Z$. In the next section, we translate this implication into conditional moment restrictions to guide model learning.

\vspace{-5mm}

\subsection{DeepGMM for CIV-DG}

Figure~\ref{fig:1} (C) illustrates the framework of the proposed CIV-DG via DeepGMM. With $R=\phi(X)$ and $p=f(R,D)$, define the residual $e:=Y-p$. Under $Z \perp U \mid D$ (and exclusion), we enforce conditional moment restrictions by making $e$ orthogonal to a rich class of centered test functions of $Z$ (and $D$), optimized with alternating critic--primal updates.

\noindent\textbf{From CIV to Conditional Moment Restrictions.} To disentangle pathology from site effects, we require the prediction residual $e = Y - f_\theta(X, D)$ to be conditionally mean-independent of instrument-induced variation. We include $D$ in the predictor to explicitly absorb demographic case-mix. Under correct specification of the predictor and exclusion restrictions, the residual captures unmodeled components correlated with artifacts; CIV implies no residual--instrument dependence conditional on $D$.

We rely on three structural assumptions: (1) \textbf{Relevance}: $Z$ induces nontrivial variation in artifact-related features; (2) \textbf{Exclusion}: $Z$ does not directly cause the diagnostic label $Y$ (except via $X$); and (3) \textbf{CIV}: $Z \perp U \mid D$. This implies that within demographic strata, the conditional mean of $e$ should not vary with $Z$, i.e., $\mathbb{E}[e \mid Z, D] = \mathbb{E}[e \mid D]$.

To enforce this, we introduce a critic function $c_\omega(Z, D) \in \mathbb{R}^M$ and minimize the empirical norm of the unconditional moment derived via the Law of Iterated Expectations (LIE):
\begin{equation}
\mathbb{E}\big[e \, \tilde{c}_\omega(Z, D)^\top\big] = \mathbb{E}\Big[ \mathbb{E}[e \mid Z, D] \, \tilde{c}_\omega(Z, D)^\top \Big] = 0,
\label{eq:centered_moment}
\end{equation}
where $\tilde{c}_\omega(Z, D) = c_\omega(Z, D) - \mathbb{E}[c_\omega \mid D]$ is the centered instrument. The equality holds because $\mathbb{E}[\tilde{c}_\omega \mid D] = 0$, rendering $\tilde{c}_\omega$ orthogonal to any function solely dependent on $D$. In practice, we estimate $\mathbb{E}[c_\omega \mid D]$ using stratum-wise moving averages to stabilize training against sampling noise.

\noindent\textbf{DeepGMM Instantiation and Optimization.} We instantiate the framework as a minimax game between a predictor $f_\theta$ and a critic $c_\omega$. As shown in Fig.~\ref{fig:1}(C), the predictor $f_\theta(X, D)$ comprises a frozen encoder $\phi$, a Causal Adapter, and a classifier head, outputting logits $s = f_\theta(X, D)$ and probability vectors $p = \text{softmax}(s)$. The critic $c_\omega$ processes concatenated inputs $[Enc(Z), Enc(D)]$ via SN-Linear $\&$ LeakyReLU MLP. Here, $Enc(\cdot)$ denotes learnable embeddings for discrete metadata (hospital ID and demographics), trained jointly with the critic. The critic finally outputs a raw instrument vector $c_\omega \in \mathbb{R}^M$, which is subsequently centered to obtain $\tilde{c}_\omega = c_\omega - \mathbb{E}[c_\omega \mid D]$.

To efficiently estimate the conditional expectation $\mathbb{E}[c_\omega \mid D]$ within a mini-batch, we employ a stratum-wise Exponential Moving Average (EMA). Let $\mathcal{K} = \{1, \dots, K\}$ denote the set of discrete group indices. We maintain a running mean vector $\mu_k$ for each $k \in \mathcal{K}$. During the forward pass, the centered instrument is obtained as $\tilde{c}_i = c_\omega(z_i, d_i) - \mu_{k(d_i)}$, where $k(d_i)$ maps the sample to its group index. Crucially, the EMA update is skipped for groups not present in the current mini-batch to ensure stability.

The core training signal arises from the empirical moment matrix $\hat{m}$, defined as the outer product of the prediction residuals $e_i = y_i - p_i \in \mathbb{R}^C$ (where $y_i$ is the one-hot label) and the centered critic features:
\begin{equation}
\hat{m}(\theta, \omega) = \frac{1}{B} \sum_{i=1}^B e_i \tilde{c}_i^\top \in \mathbb{R}^{C \times M}.
\end{equation}
The optimization proceeds via alternating updates. In each iteration, we perform $n_c$ steps of gradient ascent on $\omega$ to maximize $\mathcal{L}_\omega$ (or minimize $-\mathcal{L}_\omega$), followed by one step of gradient descent on $\theta$ to minimize $\mathcal{L}_\theta$. The objectives are defined as:
\begin{equation}
\mathcal{L}_\theta = \mathcal{L}_{\text{task}} + \lambda \mathcal{L}_{\text{GMM}}, \quad \mathcal{L}_\omega = \mathcal{L}_{\text{GMM}} - \beta \Omega(\omega).
\end{equation}
Here, $\mathcal{L}_{\text{GMM}} = \|\hat{m}\|_F^2$, and $\lambda$ controls the strength of the causal constraint. Note that the use of Spectral Normalization (SN) in the critic architecture works in tandem with $\Omega(\omega)$ (e.g., weight decay) to restrict the Lipschitz constant, preventing unbounded scaling during the maximization step.

\section{Experiment \& Analysis}

\subsection{Datasets}

\noindent\textbf{Camelyon17-WILDS (Semi-Synthetic).} We utilize whole-slide images from 5 centers ($Z \in \{0, \dots, 4\}$)~\cite{20}. To verify causal disentanglement, we induce spurious correlations by subsampling such that a synthetic stratification variable $D \in \{0, 1\}$ strongly correlates with $Z$ (enriching $D=0$ in centers 0--1 and $D=1$ in 3--4), while keeping diagnostic labels $Y$ unchanged. This forces the model to ignore site-specific artifacts to recover true pathological signals.

\noindent\textbf{Multi-Source CXR (Real-World).} We combine \textit{NIH-CXR14}~\cite{22}, \textit{CheXpert}~\cite{23}, and \textit{MIMIC-CXR}~\cite{24}. The dataset source serves as the instrument $Z$, proxying acquisition shifts. Unlike the synthetic setup, we use real patient metadata (Age and Gender) as the conditioning variable $D$ to evaluate robustness across diverse populations.

\subsection{Experimental Setup}

\noindent\textbf{Backbones \& Metrics.} 
For CXR, we leverage a frozen MedCLIP (ViT-L/14) backbone adapted via a lightweight 2-layer MLP; for Camelyon17, we train a ViT-B/16 from scratch to ensure fair comparison. Model selection (including early stopping and $\lambda$ tuning) is strictly performed on source-domain validation sets to prevent data leakage. We report Accuracy, Worst-Group Accuracy (Wg-Acc), and Expected Calibration Error (ECE) for the binary Camelyon17 task (WILDS split). For the multi-label CXR task, we employ the Leave-One-Domain-Out (LODO) protocol, reporting Macro-AUROC, Equalized Odds Difference (EOD), and Demographic Parity Difference (DPD). All results are averaged over 5 independent runs.

\noindent\textbf{Implementation Details.} 
We employ the AdamW optimizer with a weight decay of $1\times 10^{-4}$ on a single NVIDIA A800 (80\,GB) GPU.
\textit{1) For CXR:} We use a batch size of $32$ and a learning rate of $1\times 10^{-3}$ (cosine schedule with $5\%$ warmup). Demographic strata are constructed by intersecting Age quartiles with Gender groups.
\textit{2) For Camelyon17:} We use a batch size of $128$ and a learning rate of $3\times 10^{-4}$, following standard OOD protocols.
\textit{3) DeepGMM Strategy:} The critic is a 3-layer SN-MLP designed to minimize moment violations. We estimate conditional moments via stratum-wise EMA (momentum $0.9$) and perform $n_c=5$ critic updates per predictor step. The penalty weight is set to $\lambda=1.0$ (selected from $\{0.1, 1.0, 10.0\}$), with $\beta=1.0$.

\subsection{Comparisons with the state-of-the-art methods}

\begin{table*}[!t]
\centering
\caption{Main Results: Comparison with State-of-the-Art Methods. All scores presented in percentage. The best results are highlighted in bold.}
\label{tab:1}
\renewcommand{\arraystretch}{1.2} 
\setlength{\tabcolsep}{3pt}
\resizebox{1.0\textwidth}{!}{%
\begin{tabular}{|l|ccc|ccc|}
\hline
\multirow{2}{*}{\textbf{Method}} & \multicolumn{3}{c|}{\textbf{Camelyon17}} & \multicolumn{3}{c|}{\textbf{CXR}} \\
\cline{2-7} 
 & \textbf{Acc} ($\uparrow$) & \textbf{Wg-Acc} ($\uparrow$) & \textbf{ECE} ($\downarrow$) & \textbf{AUC} ($\uparrow$) & \textbf{EOD} ($\downarrow$) & \textbf{DPD} ($\downarrow$) \\
\hline 
SWAD~\cite{26}              & $92.45 \pm 0.4$ & $67.82 \pm 1.2$ & $10.15 \pm 0.5$ & $82.64 \pm 0.3$ & $8.42 \pm 0.4$ & $9.15 \pm 0.4$ \\
LISA~\cite{27}              & $92.18 \pm 0.5$ & $68.94 \pm 1.1$ & $9.84 \pm 0.6$  & $82.91 \pm 0.3$ & $8.15 \pm 0.4$ & $8.86 \pm 0.5$ \\
DFR~\cite{28}               & $91.92 \pm 0.6$ & $69.85 \pm 1.0$ & $9.62 \pm 0.5$  & $83.12 \pm 0.2$ & $7.94 \pm 0.3$ & $8.53 \pm 0.4$ \\
\hline
CLIP (Zero-Shot)~\cite{29}  & $85.23 \pm 0.5$ & $52.16 \pm 1.8$ & $18.52 \pm 1.2$ & $79.54 \pm 0.0$ & $14.81 \pm 0.0$ & $16.25 \pm 0.0$ \\
MedCLIP~\cite{30}           & $84.58 \pm 0.6$ & $50.82 \pm 1.5$ & $19.24 \pm 1.1$ & $81.23 \pm 0.2$ & $11.56 \pm 0.3$ & $12.42 \pm 0.4$ \\
AdFair-CLIP~\cite{31}       & $86.12 \pm 0.4$ & $53.45 \pm 1.6$ & $17.83 \pm 1.0$ & $83.51 \pm 0.2$ & $7.24 \pm 0.3$ & $7.83 \pm 0.3$ \\
\hline
\textbf{CIV-DG} (Ours)      & $\mathbf{93.24 \pm 0.3}$ & $\mathbf{70.46 \pm 0.9}$ & $\mathbf{8.73 \pm 0.3}$ & $\mathbf{84.15 \pm 0.2}$ & $\mathbf{6.81 \pm 0.3}$ & $\mathbf{6.78 \pm 0.3}$ \\
\hline
\end{tabular}%
}
\end{table*}

Table~\ref{tab:1} reports the main results of CIV-DG in comparison with state-of-the-art Domain Generalization (DG) methods (SWAD, LISA, DFR), Vision-Language (VL) methods (CLIP, MedCLIP, AdFair-CLIP), and an ablation variant (DeepGMM w/o CIV). Overall, CIV-DG delivers the strongest and most consistent performance across OOD robustness, fairness, and calibration.

\noindent \textbf{(1) Superior OOD Robustness on Camelyon17.}
    CIV-DG achieves the best overall OOD performance on Camelyon17, obtaining the highest Accuracy (93.24$\pm$0.3\%) and the best Worst-Group Accuracy (70.46$\pm$0.9\%). Compared to the strongest DG baseline in Wg-Acc (DFR: 69.85$\pm$1.0\%), CIV-DG provides a clear improvement, indicating better robustness under distribution shifts and reduced reliance on site-specific spurious cues.

\noindent \textbf{(2) Breaking the Fairness--Utility Trade-off on CXR.}
    On the multi-source CXR fairness benchmark, CIV-DG achieves the highest diagnostic utility (AUC: 84.15$\pm$0.2\%) while simultaneously exhibiting the smallest group disparities (EOD: 6.81$\pm$0.3\%, DPD: 6.78$\pm$0.3\%). In contrast, VL baselines such as CLIP and MedCLIP show substantially larger fairness gaps (e.g., EOD 14.81\% and 11.56\%) and lower AUC, while AdFair-CLIP improves fairness but remains inferior to CIV-DG in both AUC and disparity metrics. These results suggest that CIV-DG improves equity without sacrificing predictive performance.

\noindent \textbf{(3) Improved Calibration for Reliable Deployment.}
    Beyond accuracy, CIV-DG also yields the best calibration on Camelyon17, achieving the lowest ECE of 8.73$\pm$0.3\%. This is notably better than both VL approaches (ECE $\approx$ 18--19\%) and competitive DG baselines (ECE 9.62--10.15\%), supporting that CIV-DG produces more reliable confidence estimates under OOD shifts.

\subsection{Ablation studies}

\vspace{-5mm}

\begin{table}[h]
\centering
\caption{Ablation study on CIV-DG components. All scores presented in percentage. The best results are highlighted in bold.}
\label{tab:2}
\renewcommand{\arraystretch}{1.2} 
\resizebox{\linewidth}{!}{
\begin{tabular}{|l|cc|ccc|ccc|}
\hline
\multirow{2}{*}{\textbf{Method}} & \multicolumn{2}{c|}{\textbf{Components}} & \multicolumn{3}{c|}{\textbf{Camelyon17}} & \multicolumn{3}{c|}{\textbf{CXR}} \\ \cline{2-9} 
 & \textbf{IV} & \textbf{Cond.} & \textbf{Acc} ($\uparrow$) & \textbf{Wg-Acc} ($\uparrow$) & \textbf{ECE} ($\downarrow$) & \textbf{AUC} ($\uparrow$) & \textbf{EOD} ($\downarrow$) & \textbf{DPD} ($\downarrow$) \\ \hline
ERM (Baseline)       & $-$ & $-$ & $88.50 \pm 0.6$ & $55.20 \pm 1.5$ & $15.42 \pm 0.8$ & $80.12 \pm 0.4$ & $12.45 \pm 0.5$ & $13.10 \pm 0.5$ \\
DeepGMM (w/o CIV)    & \checkmark & $-$ & $91.25 \pm 0.5$ & $63.54 \pm 1.2$ & $11.82 \pm 0.6$ & $81.43 \pm 0.3$ & $10.12 \pm 0.4$ & $10.85 \pm 0.5$ \\
CIV-DG (Random $Z$)  & \checkmark & Rand & $92.10 \pm 0.4$ & $66.85 \pm 1.1$ & $10.05 \pm 0.5$ & $82.50 \pm 0.3$ & $8.95 \pm 0.4$ & $9.20 \pm 0.4$ \\ \hline
\textbf{CIV-DG}      & \checkmark & \textbf{Causal} & $\mathbf{93.24 \pm 0.3}$ & $\mathbf{70.46 \pm 0.9}$ & $\mathbf{8.73 \pm 0.3}$ & $\mathbf{84.15 \pm 0.2}$ & $\mathbf{6.81 \pm 0.3}$ & $\mathbf{6.78 \pm 0.3}$ \\ \hline
\end{tabular}
}
\end{table}

Table~\ref{tab:2} validates the efficacy of our conditional instrumentation. Compared to the DeepGMM baseline, the full CIV-DG improves Camelyon17 Wg-Acc by 6.92$\%$ ($63.54{\rightarrow}70.46$) and CXR AUC by 2.72$\%$ ($81.43{\rightarrow}84.15$), while reducing fairness gaps (EOD/DPD) by over 3 points. Furthermore, CIV-DG significantly outperforms the Random Z variant (66.85$\%$ Wg-Acc), confirming that performance gains stem from meaningful causal guidance rather than stochastic noise or increased model capacity.

\subsection{Visualization and Discussion}

\begin{figure*}[!t]
    \centering
    \includegraphics[width=0.99\linewidth]{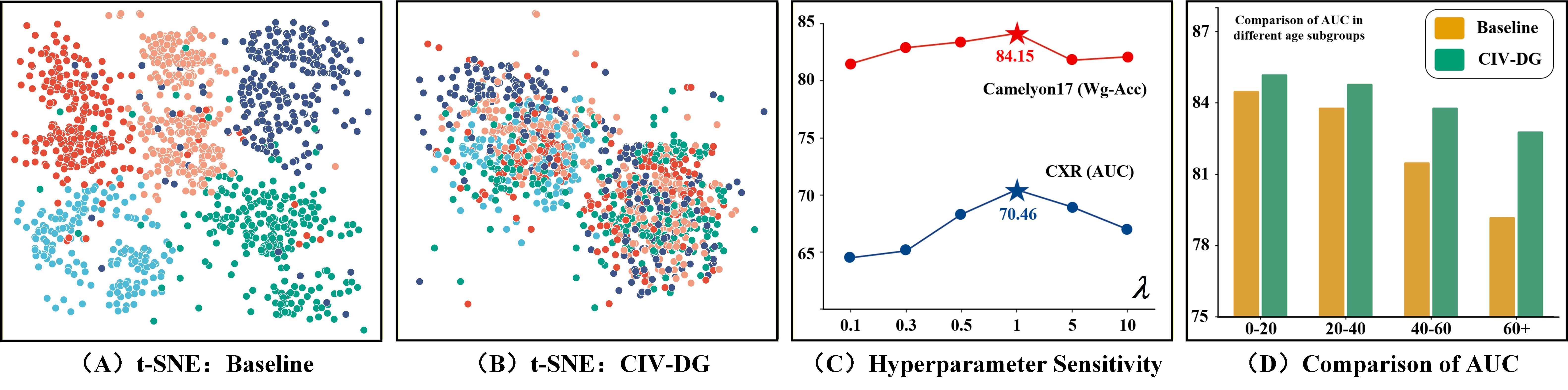}
    \caption{\textbf{Qualitative analysis of CIV-DG. (A-B)} t-SNE visualization showing improved feature separation over Baseline. \textbf{(C)} Hyperparameter sensitivity analysis of $\lambda$. \textbf{(D)} AUC comparison across age subgroups on CXR.}
    \label{fig:2}
\end{figure*}

To further understand the effectiveness of CIV-DG, we provide a comprehensive analysis in Figure~\ref{fig:2}. \textbf{(1)} To verify whether the model relies on hospital-specific styles (spurious correlations), we visualize the features colored by hospital IDs through t-SNE on Camelyon17. In Figure~\ref{fig:2} (A) (Baseline), samples from different hospitals form distinct, isolated clusters (e.g., the red and blue clusters are clearly separated). This indicates that the baseline model primarily learns domain-specific biases rather than true pathological features, leading to poor generalization. In contrast, Figure~\ref{fig:2} (B) (CIV-DG) shows a uniform mixture of samples from all hospitals in two separate clusters (yes/no lesions), proving that CIV-DG successfully removes domain-specific artifacts and learns a truly domain-invariant representation. \textbf{(2)} Regarding hyperparameter sensitivity (Figure~\ref{fig:2} (C)), we evaluate the impact of the weight coefficient $\lambda$ on both datasets. The performance trend shows an initial increase followed by stability, peaking at $\lambda=1.0$. The results demonstrate that CIV-DG is robust to hyperparameter variations within a reasonable range. \textbf{(3)} Finally, regarding subgroup robustness (Figure~\ref{fig:2} (D)), we compare the AUC performance across different age groups on the CXR dataset. While the baseline suffers a severe performance drop in the elderly group (60+), CIV-DG consistently outperforms the baseline across all subgroups. Notably, our method significantly mitigates the accuracy degradation in the hardest subgroup (60+), highlighting its superior generalization capability and fairness across demographics.

\section{Conclusion}

In this paper, we identified selection bias driven by non-random patient assignment as a fundamental barrier to medical AI generalization. To surmount this, we proposed \textbf{CIV-DG}, a causal framework leveraging Conditional Instrumental Variables to disentangle pathology from site-specific artifacts. Unlike standard methods assuming random assignment, CIV-DG enforces conditional moment restrictions via a DeepGMM architecture to handle endogenous hospital selection. Extensive experiments on Camelyon17 and multi-source CXR datasets demonstrate that CIV-DG achieves state-of-the-art performance in both OOD robustness and fairness, significantly outperforming strong baselines. These results validate that modeling the causal structure of data acquisition is essential for reliable clinical deployment. Future work will extend this framework to multi-modal records and continuous confounders.

\clearpage
%
%
%

\begin{thebibliography}{}

\bibitem{1}
Moor M, Banerjee O, Abad Z S H, et al. Foundation models for generalist medical artificial intelligence[J]. Nature, 2023, 616(7956): 259-265, \url{https://www.nature.com/articles/s41586-023-05881-4}

\bibitem{2}
Thirunavukarasu A J, Ting D S J, Elangovan K, et al. Large language models in medicine[J]. Nature medicine, 2023, 29(8): 1930-1940, \url{https://www.nature.com/articles/s41591-023-02448-8}

\bibitem{3}
Xiang J, Wang X, Zhang X, et al. A vision–language foundation model for precision oncology[J]. Nature, 2025, 638(8051): 769-778, \url{https://www.nature.com/articles/s41586-024-08378-w}

\bibitem{4}
Ouyang C, Chen C, Li S, et al. Causality-inspired single-source domain generalization for medical image segmentation[J]. IEEE Transactions on Medical Imaging, 2022, 42(4): 1095-1106, \url{https://ieeexplore.ieee.org/abstract/document/9961940}

\bibitem{5}
Zhou K, Liu Z, Qiao Y, et al. Domain generalization: A survey[J]. IEEE transactions on pattern analysis and machine intelligence, 2023, 45(4): 4396-4415, \url{https://ieeexplore.ieee.org/abstract/document/9847099}

\bibitem{6}
Acosta J N, Falcone G J, Rajpurkar P, et al. Multimodal biomedical AI[J]. Nature medicine, 2022, 28(9): 1773-1784, \url{https://www.nature.com/articles/s41591-022-01981-2}

\bibitem{8}
Seyyed-Kalantari L, Zhang H, McDermott M B A, et al. Underdiagnosis bias of artificial intelligence algorithms applied to chest radiographs in under-served patient populations[J]. Nature medicine, 2021, 27(12): 2176-2182, \url{https://www.nature.com/articles/s41591-021-01595-0}

\bibitem{9}
Geirhos R, Jacobsen J H, Michaelis C, et al. Shortcut learning in deep neural networks[J]. Nature Machine Intelligence, 2020, 2(11): 665-673, \url{https://www.nature.com/articles/s42256-020-00257-z}

\bibitem{10}
Zhou W, Liu F, Zheng H, et al. Mitigating data bias and ensuring reliable evaluation of AI models with shortcut hull learning[J]. Nature Communications, 2025, 16(1): 5513, \url{https://www.nature.com/articles/s41467-025-60801-6}

\bibitem{11}
Ricci Lara M A, Echeveste R, Ferrante E. Addressing fairness in artificial intelligence for medical imaging[J]. nature communications, 2022, 13(1): 4581, \url{https://www.nature.com/articles/s41467-022-32186-3}

\bibitem{12}
Mohyuddin G R, Prasad V. Detecting selection bias in observational studies—when interventions work too fast[J]. JAMA internal medicine, 2023, 183(9): 897-898, \url{https://jamanetwork.com/journals/jamainternalmedicine/article-abstract/2805974}

\bibitem{13}
Castro D C, Walker I, Glocker B. Causality matters in medical imaging[J]. Nature Communications, 2020, 11(1): 3673, \url{https://www.nature.com/articles/s41467-020-17478-w}

\bibitem{14}
McLeod G A, Stanley E A M, Rosenal T, et al. Distinct visual biases affect humans and artificial intelligence in medical imaging diagnoses[J]. npj Digital Medicine, 2025, \url{https://www.nature.com/articles/s41746-025-02226-5}

\bibitem{16}
Angrist J D, Imbens G W, Rubin D B. Identification of causal effects using instrumental variables[J]. Journal of the American statistical Association, 1996, 91(434): 444-455, \url{https://www.tandfonline.com/doi/abs/10.1080/01621459.1996.10476902}

\bibitem{17}
Wu A, Kuang K, Xiong R, et al. Instrumental variables in causal inference and machine learning: A survey[J]. ACM Computing Surveys, 2025, 57(11): 1-36, \url{https://dl.acm.org/doi/full/10.1145/3735969}

\bibitem{18}
Sun Q, Xia H, Liu J. Data-faithful feature attribution: Mitigating unobservable confounders via instrumental variables[J]. Advances in Neural Information Processing Systems, 2024, 37: 44935-44964.

\bibitem{19}
Bennett A, Kallus N, Schnabel T. Deep generalized method of moments for instrumental variable analysis[J]. Advances in neural information processing systems, 2019, 32, \url{https://proceedings.neurips.cc/paper/2019/hash/15d185eaa7c954e77f5343d941e25fbd-Abstract.html}

\bibitem{20}
Koh P W, Sagawa S, et al. Wilds: A benchmark of in-the-wild distribution shifts[C]//International conference on machine learning. PMLR, 2021: 5637-5664, \url{https://par.nsf.gov/biblio/10300276}

\bibitem{21}
Sagawa S, Koh P W, Hashimoto T B, et al. Distributionally robust neural networks for group shifts: On the importance of regularization for worst-case generalization[J]. arXiv preprint arXiv:1911.08731, 2019, \url{https://arxiv.org/abs/1911.08731}

\bibitem{22}
Wang X, Peng Y, Lu L, et al. Chestx-ray8: Hospital-scale chest x-ray database and benchmarks on weakly-supervised classification and localization of common thorax diseases[C]//Proceedings of the IEEE conference on computer vision and pattern recognition. 2017: 2097-2106, \url{https://openaccess.thecvf.com/content_cvpr_2017/html/Wang_ChestX-ray8_Hospital-Scale_Chest_CVPR_2017_paper.html}

\bibitem{23}
Irvin J, Rajpurkar P, Ko M, et al. Chexpert: A large chest radiograph dataset with uncertainty labels and expert comparison[C]//Proceedings of the AAAI conference on artificial intelligence. 2019, 33(01): 590-597, \url{https://ojs.aaai.org/index.php/AAAI/article/view/3834}

\bibitem{24}
Johnson A E W, Pollard T J, Berkowitz S J, et al. MIMIC-CXR, a de-identified publicly available database of chest radiographs with free-text reports[J]. Scientific data, 2019, 6(1): 317, \url{https://www.nature.com/articles/s41597-019-0322-0}

\bibitem{26}
Cha J, Chun S, Lee K, et al. Swad: Domain generalization by seeking flat minima[J]. Advances in Neural Information Processing Systems, 2021, 34: 22405-22418, \url{https://proceedings.neurips.cc/paper_files/paper/2021/hash/bcb41ccdc4363c6848a1d760f26c28a0-Abstract.html}

\bibitem{27}
Yao H, Wang Y, Li S, et al. Improving out-of-distribution robustness via selective augmentation[C]//International Conference on Machine Learning. PMLR, 2022: 25407-25437, \url{https://proceedings.mlr.press/v162/yao22b.html}

\bibitem{28}
Kirichenko P, Izmailov P, Gordon Wilson A. Last Layer Re-Training is Sufficient for Robustness to Spurious Correlations[J]. ICLR 2023, 2023, \url{https://par.nsf.gov/biblio/10437678}

\bibitem{29}
Radford A, Kim J W, Hallacy C, et al. Learning transferable visual models from natural language supervision[C]//International conference on machine learning. PmLR, 2021: 8748-8763, \url{https://proceedings.mlr.press/v139/radford21a}

\bibitem{30}
Wang Z, Wu Z, Agarwal D, et al. Medclip: Contrastive learning from unpaired medical images and text[C]//Proceedings of the 2022 Conference on Empirical Methods in Natural Language Processing. 2022: 3876-3887, \url{https://aclanthology.org/2022.emnlp-main.256/}

\bibitem{31}
Yi C, Xiong Z, Qi Q, et al. AdFair-CLIP: Adversarial Fair Contrastive Language-Image Pre-training for Chest X-rays[C]//International Conference on Medical Image Computing and Computer-Assisted Intervention. Cham: Springer Nature Switzerland, 2025: 13-23, \url{https://link.springer.com/chapter/10.1007/978-3-032-04978-0_2}








\end{thebibliography}
%

\end{document}